\pdfoutput=1
\documentclass[10pt,twocolumn,letterpaper]{article}
\usepackage[pagenumbers]{cvpr}
\usepackage{graphicx}
\usepackage{amsmath,amssymb}
\usepackage{booktabs,multirow,array}
\usepackage{xcolor}
\usepackage{tikz}
\usetikzlibrary{arrows.meta,positioning,fit}
\newcommand{\method}{CoachSI}
\newcommand{\policy}{\pi_{\theta}}
\newcommand{\coach}{c_{\phi}}
\newcommand{\success}{\mathrm{succ}}

\definecolor{cvprblue}{rgb}{0.21,0.49,0.74}
\usepackage[pagebackref,breaklinks,colorlinks,allcolors=cvprblue]{hyperref}
\title{CoachSI: Self-Improvement of Robot Policies\\with a Language Coach}
\author{Jiachen Tao \quad Gengyu Zhang \quad Haoxuan Wang \quad Bing Liu \quad Yan Yan\textsuperscript{\dag}\\[2pt]
University of Illinois at Chicago\\[2pt]
{\small\url{https://tlxhlll.github.io/CoachSI/}}}
\begin{document}
\twocolumn[{\maketitle\begin{center}
\captionsetup{hypcap=false}
\includegraphics[width=\textwidth]{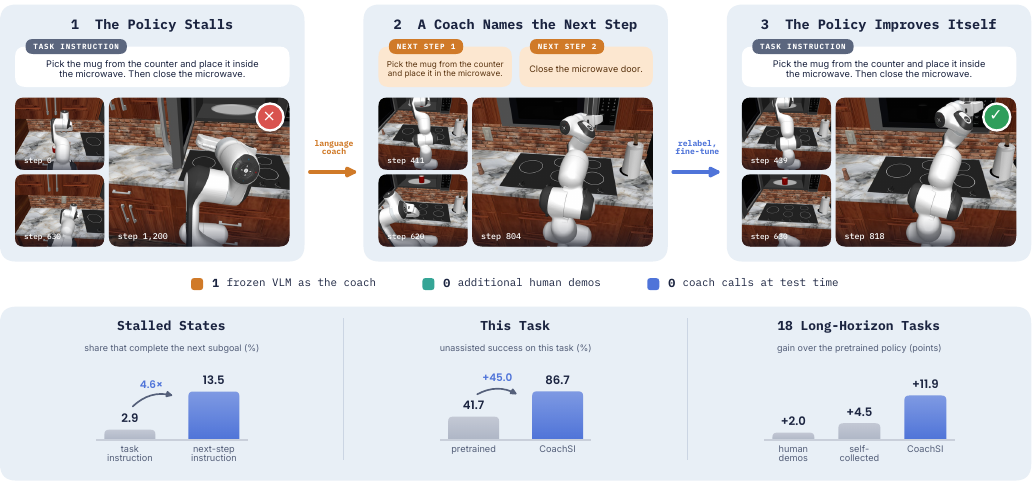}
\captionof{figure}{\textbf{\method{}: self-improvement with a language coach.} \emph{(1)} Given the task instruction, the pretrained policy picks up the mug and stalls; it does not complete any subgoal in 1,800 steps. \emph{(2)} A frozen VLM coach reads the camera images and names the next step; the same policy, from the same initial state, completes the task. \emph{(3)} Fine-tuned on its coached successes relabeled with the task instruction, the policy completes the task from the task instruction alone, with no coach at test time. \emph{Results:} across 500 stalled states from two policies, the correct next-step instruction raises subgoal completion by $4.6\times$, and a VLM coach recovers 91\% of this gain from images alone. \method{} raises unassisted success from 41.7\% to 86.7\% on this task and from 47.4\% to 59.3\% across 18 long-horizon tasks, a gain of 11.9 points where human demonstrations and self-collected successes add 2.0 and 4.5.}
\label{fig:teaser}
\end{center}
}]
{\let\thefootnote\relax\footnotetext{\textsuperscript{\dag}Corresponding author.}}
\begin{abstract}
Generalist robot policies execute atomic skills reliably but fail when the same skills are composed into long-horizon tasks. Self-improvement from the policy's own experience does not close this gap: experience collected by the policy alone stays within its own distribution and reproduces what it can already do. We show that what the policy lacks is not a skill but the next step. We present \method{}, a self-improvement method in which a frozen, off-the-shelf vision-language model (VLM) coaches the policy in language: it watches the robot and names the next step in the format of the policy's atomic instructions, and never outputs an action. Successful coached rollouts are relabeled in hindsight with the original task instruction and used to fine-tune the policy. The fine-tuned policy is then deployed on its own. On states where a World Action Model and a Vision-Language-Action model have stalled, naming the next step raises subgoal completion by $4.6\times$ over the unchanged task instruction, and the frozen VLM supplies it from the same camera images the policy sees. After fine-tuning, unassisted success on long-horizon tasks in unseen scenes rises from 47.4\% to 59.3\%, 9.8 points above fine-tuning on human demonstrations and 7.3 points above fine-tuning on self-collected successes at an equal data budget. With a coach that sees only camera images, it reaches 56.8\%, above both baselines. The gain is obtained with the action backbone frozen: language coaching teaches the policy what should happen next, and the skills it already has do the rest.
\end{abstract}

\section{Introduction}
\label{sec:intro}
Humans become skilled in the physical world not by being shown every motion, but by practice: they act, observe the outcome, and adjust. Repeated over many attempts, this loop turns a clumsy first try into a reliable skill, and it continues long after instruction ends. Generalist robot policies now perform a wide range of manipulation tasks~\cite{brohan2023rt1,brohan2023rt2,oxe2024,octo2024,kim2024openvla,black2025pi0,black2025pi05,bjorck2025gr00t}, yet they rarely improve this way: when a policy fails, making it better typically means collecting new human demonstrations. Enabling a robot to improve from its own experience is the goal of self-improvement in physical AI~\cite{luo2026rsi,bousmalis2023robocat}, and it would scale robot learning beyond what human teleoperation can supply.

The most direct form of self-improvement changes the policy itself: the robot acts, keeps its successful rollouts, and is fine-tuned on them. This recipe has two limitations. First, a policy acting alone stays within its own distribution, so its successes show it only what it can already do. Second, on the long-horizon tasks where improvement matters most, the policy fails often and has few successes to collect. Experience improves a policy when it contains successes the policy cannot yet produce, and self-collected experience contains few of them.

Existing methods obtain such successes with supervision from outside the policy. Interactive imitation learning brings in a human operator who takes over when the robot errs and supplies corrective actions~\cite{kelly2019hgdagger,amin2026pistar06}. Every correction costs operator time, so the supervision scales with human effort, the cost that self-improvement is meant to remove. Reinforcement learning replaces the operator with reward and value models that score the robot's own attempts~\cite{ghasemipour2025selfimproving,chen2026sarm2}. These models must themselves be trained before they can supervise the policy. Hierarchical systems place a vision-language model above the policy to decompose the task into steps~\cite{ahn2022saycan,shi2025hirobot}. The policy itself is left unchanged, and the system depends on the high-level model at every deployment.

To understand what keeps a policy from producing these successes on its own, we analyze rollouts of generalist policies on long-horizon tasks. We find that the missing successes are closer than they appear. The policies execute atomic skills reliably, yet fail when the same skills are composed into a long-horizon task~\cite{mees2022calvin,liu2023libero,fan2025longvla,tao2026prism}, and the failures take a specific form: partway through the task the policy stalls, and a stalled policy almost never recovers on its own. In a recurring case, the robot is already at the right object and holds the required skill, yet does not act. \textbf{What the policy lacks in that moment is not a skill but the next step.} What it needs is therefore far less than corrective actions or rewards: a single instruction.

We introduce \method{} (\cref{fig:teaser}), a self-improvement method guided by language alone. Its central component is a language coach: a frozen, off-the-shelf vision-language model (VLM) that watches the robot and replaces the task instruction with the next step, phrased in the format of the policy's atomic instructions. The coach never outputs an action; every motion is produced by the policy's own skills. Successful coached rollouts are relabeled in hindsight with the original task instruction~\cite{andrychowicz2017her} and used to fine-tune the policy, which is then deployed on its own. On a World Action Model, fine-tuning updates only the video backbone and keeps the action backbone frozen. \method{} needs no human operator, no reward model, and no VLM at test time.

We evaluate \method{} on RoboCasa365~\cite{nasiriany2026robocasa365} with a World Action Model (WAM)~\cite{wang2026openwam} and a Vision-Language-Action model (VLA)~\cite{xiaomi2026robotics}. On states where the two policies have stalled, naming the next step raises subgoal completion by $4.6\times$ over the unchanged task instruction, and a VLM coach that sees only the policy's camera images delivers 91\% of the gain of an oracle that always names the correct step. With the coach, the policy completes more than a quarter of the episodes it fails on its own. After fine-tuning the WAM on coached experience, unassisted success on 18 long-horizon tasks in unseen scenes rises from 47.4\% to 59.3\%, 9.8 points above fine-tuning on human demonstrations and 7.3 points above fine-tuning on self-collected successes at an equal data budget. With a coach that sees only camera images, it reaches 56.8\%, above both baselines. The action backbone is frozen throughout, so the entire gain comes from what the policy predicts will happen next: the skills were already there.

Our contributions are:
\begin{itemize}
    \item We identify a class of stalls in long-horizon manipulation in which the policy has the skill but lacks the next step, and show that a single next-step instruction resolves them, for both a WAM and a VLA.
    \item We propose \method{}, a self-improvement method of language coaching, hindsight relabeling, and fine-tuning, guided only by a frozen VLM's next-step instruction.
    \item We show that \method{} turns coached experience into unassisted capability and outperforms both human demonstrations and self-collected experience at an equal data budget.
\end{itemize}

\section{Related Work}
\label{sec:related}
\paragraph{Self-improving robot policies.}
A growing body of work improves robot policies with experience collected after pretraining~\cite{luo2026rsi}. Autonomous-improvement systems let the policy practice and train on its own rollouts after hindsight relabeling or success filtering~\cite{bousmalis2023robocat,zhou2024soar}. Reinforcement-learning methods learn reward or value functions and optimize the policy against them~\cite{ghasemipour2025selfimproving,xiao2025pld,xu2025rldg,chen2026sarm2,amin2026pistar06,yang2026rise}, with VLMs increasingly serving as the reward model~\cite{ma2025gvl,lee2026roboreward}. Interactive imitation learning queries an expert, typically a human teleoperator, for corrective actions on the states the policy visits~\cite{ross2011dagger,kelly2019hgdagger,liu2023sirius,luo2024hilserl,amin2026pistar06}. Most recently, agents built on LLMs and world models improve the system around the policy: RoboRSI~\cite{wen2026roborsi} writes and repairs skill code, and RoboCoach~\cite{liu2026robocoach} selects which subtask demonstrations to collect next. In contrast to these works, \method{} improves the policy with a single signal, the next-step instruction of a frozen VLM. It trains no reward or value model, requires no action-level expert, and updates the policy itself.

\paragraph{Language guidance at deployment.}
Language is the standard interface between high-level reasoning and low-level control. Hierarchical systems decompose a task into language steps at run time, with a separate LLM or VLM planner~\cite{huang2022zeroshot,ahn2022saycan,huang2022inner,liang2023cap,shi2025hirobot,hu2026orchestrating} or inside the policy~\cite{belkhale2024rth,zawalski2024ecot,black2025pi05,yang2025lohovla}. A second line corrects the policy in language during execution, with corrections issued by humans~\cite{shi2024yell,liu2023olaf} or by VLMs that detect and explain failures~\cite{liu2023reflect,duan2024aha} or choose among the policy's candidate action plans~\cite{wu2025forewarn}. RACER~\cite{dai2024racer} trains both a VLM supervisor and a language-conditioned policy on expert-corrected failure data and keeps the supervisor in the loop at deployment. These systems rely on a human or a high-level model at test time, and several train the policy to follow a new language interface. \method{} uses the VLM only during data collection: the coach is frozen and off-the-shelf, it reuses the instruction format of the policy's atomic skills, and the fine-tuned policy executes the task instruction on its own.

\paragraph{Language as a training signal.}
Language also serves as a signal for training. Hindsight relabeling assigns to a trajectory the goal it reached~\cite{andrychowicz2017her}, and DIAL~\cite{xiao2023dial} and SOAR~\cite{zhou2024soar} use VLMs to produce such instruction labels for offline or autonomously collected data. BOSS~\cite{zhang2023boss} has an LLM propose which skills to chain during practice and learns the chains as new skills. Scaling Up and Distilling Down~\cite{ha2023scalingup} and Manipulate-Anything~\cite{duan2024manipulate} have an LLM or a VLM generate demonstrations and distill them into a policy. In these works, language labels experience after the fact, selects what the policy practices, or drives a separate data generator. In \method{}, language intervenes at the point of failure: the coach names the next step to a policy that has stalled, in the format of the policy's own atomic instructions, so every action in the resulting experience is produced by skills the policy already has. This design lets us measure what language alone contributes, on fixed stalled states of a WAM~\cite{wang2026openwam} and a VLA~\cite{xiaomi2026robotics} in RoboCasa365~\cite{nasiriany2026robocasa365} and against the policy's own successes at an equal data budget.

\paragraph{World Action Models.}
A growing family of policies predicts future observations together with actions. UniPi~\cite{du2023unipi} casts control as text-conditioned video generation and extracts actions from the generated video. GR-1~\cite{wu2024gr1} pretrains a transformer on video generation and fine-tunes it to output future frames and actions. Video Prediction Policy~\cite{hu2025vpp} conditions its action head on the internal representations of a video diffusion model, CoT-VLA~\cite{zhao2025cotvla} predicts a subgoal image before each action chunk, the Unified Video Action model~\cite{li2025uva} learns a joint video-action latent, and Unified World Models~\cite{zhu2025uwm} couple video and action diffusion in one model. DreamZero~\cite{ye2026dreamzero} and OpenWAM~\cite{wang2026openwam} scale this design into World Action Models (WAMs) built on pretrained video-generation backbones, and Cosmos Policy~\cite{kim2026cosmospolicy} fine-tunes a pretrained video model directly into a policy. In a WAM, the video backbone predicts what will happen and the action backbone turns the prediction into motor commands. \method{} exploits this separation: coached experience updates only the video backbone, and the action backbone stays frozen.

\section{CoachSI}
\label{sec:method}
\begin{figure*}[t]
\centering
\includegraphics[width=\textwidth]{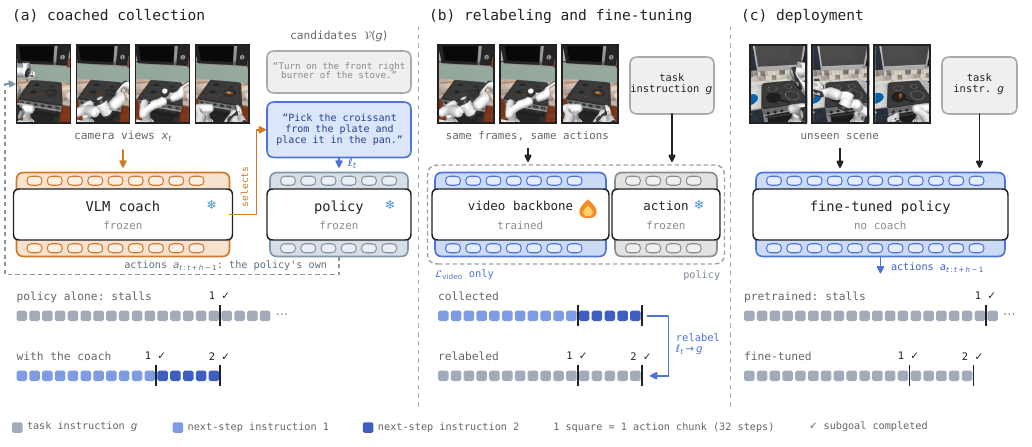}
\caption{\textbf{Overview of \method{}.} (a) \emph{Coached collection.} At every action chunk, a frozen VLM coach reads the camera views and selects the policy's instruction $\ell_t$ from the next-step candidates of the task, or keeps the task instruction $g$. The policy weights are fixed and every action is the policy's own. (b) \emph{Hindsight relabeling and fine-tuning.} Successful rollouts keep their observations and actions, and $\ell_t$ is replaced by $g$. Fine-tuning mixes them with human demonstrations and updates only the video backbone. (c) \emph{Deployment.} The fine-tuned policy receives observations and $g$ alone, with no coach. Each square is one action chunk, colored by the instruction the policy receives; ticks mark completed subgoals. The example is one long-horizon task: alone, the pretrained policy completes the first subgoal and stalls; with the coach, it completes the task; after fine-tuning, it completes the task in an unseen scene without the coach.}
\vspace{-10pt}
\label{fig:overview}
\end{figure*}

\subsection{Preliminaries}
We consider a language-conditioned policy $\policy$~\cite{jang2022bcz,brohan2023rt1} that maps an observation $o_t$ and an instruction $\ell_t$ to an action chunk~\cite{zhao2023act,chi2023diffusion} of length $h$,
\begin{equation}
 a_{t:t+h-1}\sim\policy(\cdot\mid o_t,\ell_t).
\end{equation}
The policy is trained on atomic tasks, each a single skill such as opening a cabinet, and on composite tasks that chain several skills toward one goal. A composite task is specified by a task instruction $g$. In unassisted execution, the policy is conditioned on $\ell_t=g$ for the whole episode.

A World Action Model (WAM) implements $\policy$ with two coupled components. A video backbone predicts future frames from the observation and the instruction, and an action backbone predicts the action chunk while attending to the features of the video backbone. The two are denoised jointly and trained with $\mathcal{L}=\lambda_v\mathcal{L}_{\mathrm{video}}+\lambda_a\mathcal{L}_{\mathrm{action}}$. A vision-language-action model (VLA) maps observations and language to actions without a video-prediction branch. We fine-tune the WAM, whose separate backbones let us update what the policy predicts next while every action is still produced by the pretrained weights (\cref{sec:learning}).

\subsection{Language Coach}
\label{sec:coach}
Each composite task $g$ has a set of next-step instructions $\mathcal{V}(g)=\{v_1,\ldots,v_m\}$, one per step of the task, written in the format of the policy's atomic instructions. For the task ``Pick the croissant and place it on the pan. Then turn on the stove to warm the croissant.'', $\mathcal{V}(g)$ holds ``Pick the croissant from the plate and place it in the pan.'' and ``Turn on the front right burner of the stove.'' Each $v_i$ asks for a skill the policy already has.

The coach $\coach$ is a frozen, off-the-shelf VLM. At every action-chunk boundary it receives the task instruction, the candidate set, and the camera views the policy sees, denoted $x_t$, and selects the first step that is not yet done, or returns \texttt{KEEP} when every step is done:
\begin{align}
 z_t &= \coach(g,\mathcal{V}(g),x_t),\\
 \ell_t &=
 \begin{cases}
  g, & z_t=\texttt{KEEP},\\
  v_{z_t}, & \text{otherwise}.
 \end{cases}
\end{align}
The selected instruction conditions the policy for the next action chunk, and the coach is queried again afterwards. The coach never outputs an action: every motion in a coached rollout is produced by $\policy$. We also use a state-based variant of the coach, which reads a text description of object and robot state in place of the images.

\subsection{Hindsight Relabeling and Fine-Tuning}
\label{sec:learning}
We run the coached policy from the initial state on a set of collection scenes and keep the rollouts that reach task success within the task horizon. For each kept rollout $\tau$, we retain the observations and the executed actions and replace every coaching instruction with the task instruction $g$~\cite{andrychowicz2017her}:
\begin{equation}
 \mathcal{D}_{\mathrm{coach}}=\{(\tau,g):\success(\tau,g)=1\}.
\end{equation}
The relabeled data shows the task being completed under $g$ alone.

We fine-tune the policy on $\mathcal{D}_{\mathrm{coach}}$ mixed with the RoboCasa365 human demonstrations $\mathcal{D}_{\mathrm{human}}$ on which it was trained, with a fraction $\alpha$ of the training samples drawn from $\mathcal{D}_{\mathrm{coach}}$. Because the coach outputs no actions, the actions in $\mathcal{D}_{\mathrm{coach}}$ are the policy's own and carry no new action knowledge. What is new is the future that unfolds under $g$, which is a target for video prediction. We therefore update only the self-attention layers of the video backbone, with parameters $\theta_v$, using the video loss, and keep the action backbone frozen:
\begin{equation}
 \min_{\theta_v}\;(1-\alpha)\,
 \mathbb{E}_{\mathcal{D}_{\mathrm{human}}}\big[\mathcal{L}_{\mathrm{video}}\big]
 +\alpha\,\mathbb{E}_{\mathcal{D}_{\mathrm{coach}}}\big[\mathcal{L}_{\mathrm{video}}\big].
\end{equation}

At deployment the fine-tuned policy receives $g$ for the whole episode. No coach, no candidate set, and no instruction switch is used.

\section{Experiments}
\label{sec:experiments}
\begin{figure*}[t]
\centering
\includegraphics[width=\textwidth]{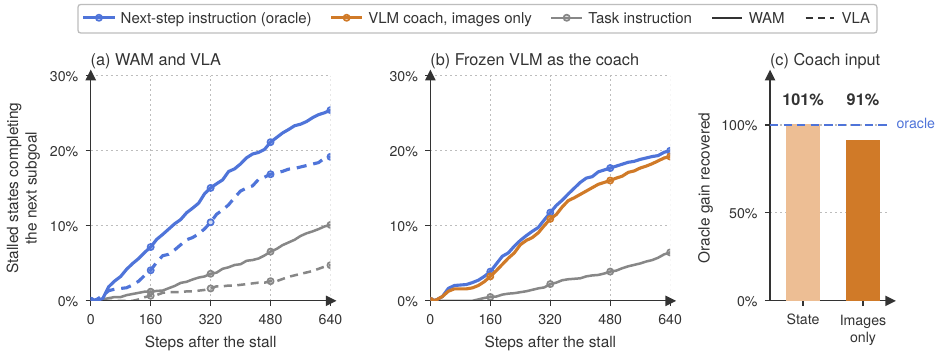}
\caption{\textbf{A single next-step instruction resolves the stall, and a frozen VLM supplies it.} (a) Share of stalled states in which the policy completes the next subgoal, as a function of steps after the stall, for a WAM and a VLA (500 states, 15 tasks). (b) On the states evaluated with every instruction source (261 states), a VLM coach that sees only camera images matches the oracle. (c) Share of the oracle gain recovered by the VLM coach at 320 steps.}
\vspace{-10pt}
\label{fig:stall}
\end{figure*}

Our experiments follow the argument of the paper. We show that generalist policies possess the skills for long-horizon tasks yet stall and that a single next-step instruction resolves the stall (\cref{sec:exp_nextstep}), that a frozen VLM supplies this instruction from camera images (\cref{sec:exp_vlm}), and that fine-tuning on coached experience turns it into the policy's own capability (\cref{sec:exp_training}).

\subsection{Experimental Setup}
\label{sec:exp_setup}

\textbf{Benchmark and policies.} We evaluate on RoboCasa365~\cite{nasiriany2026robocasa365}, which builds on RoboCasa~\cite{nasiriany2024robocasa}. The benchmark provides atomic tasks, each a single skill, and composite tasks that chain several of them; we use 18 composite tasks as our long-horizon tasks. We study two public checkpoints as released: OpenWAM~\cite{wang2026openwam}, a WAM, and Xiaomi-Robotics-1~\cite{xiaomi2026robotics}, a VLA. Data collection and evaluation use disjoint sets of kitchen scenes.

\noindent\textbf{Stalled states.} A rollout under the task instruction stalls when no subgoal changes status for 320 steps. We collect 500 stalled states from failed rollouts of the two policies on 15 tasks. From each state we restore the simulator and continue the same policy under each instruction source, with six continuations per source and matched random seeds. We report \emph{subgoal completion}, the share of continuations that complete the next subgoal within 320 steps, and \emph{task success} within 640 steps.

\noindent\textbf{Coach.} The coach is a frozen Qwen3.8-27B~\cite{qwen2026qwen38}. At every action chunk it receives three camera views and the candidate next-step instructions of the task, and returns one candidate or \texttt{KEEP} (\cref{sec:coach}). The \emph{oracle} replaces the VLM with a rule that reads subgoal status from the simulator and always selects the first unfinished step; it is the upper bound of a coach. We collect coached data with the image-only coach and with the state-based coach (\cref{sec:coach}) and report \method{} for both (\cref{tab:training}); the analyses of \cref{sec:exp_training} use the state-based coach.

\noindent\textbf{Fine-tuning and evaluation.} All fine-tuned policies start from the OpenWAM checkpoint and share one recipe: 3{,}000 gradient steps at batch size 16, with only the self-attention layers of the video backbone updated (\cref{sec:learning}). Of the 48{,}000 training samples, 80\% are RoboCasa365 human demonstrations, identical across policies. The remaining 20\% is the only variable: each of the 18 tasks receives the same budget of 530 samples, filled from human demonstrations, from the policy's own successes under the task instruction, or from coached successes. We evaluate every policy under the task instruction alone, on 20 unseen scenes with three trials per task, for 1{,}080 episodes. Comparisons are paired by episode, and 95\% confidence intervals are bootstrapped over tasks. A second training seed confirms the comparison: pooled over both seeds, \method{} exceeds human demonstrations by 10.0 points and self-collected successes by 5.5.

\begin{figure}[t]
\centering
\includegraphics[width=\columnwidth]{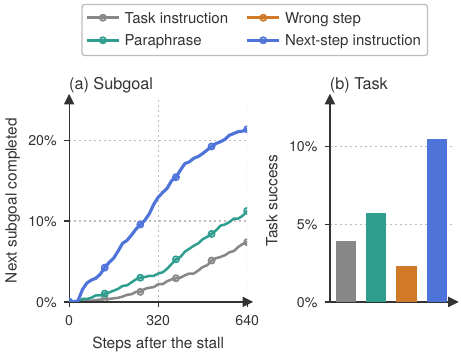}
\caption{\textbf{Instruction-content controls} on 228 stalled states of the WAM and the VLA. (a) Share of continuations that complete the next subgoal, by steps after the stall. (b) Task success within 640 steps.}
\vspace{-10pt}
\label{fig:controls}
\end{figure}

\subsection{One Next-Step Instruction Resolves the Stall}
\label{sec:exp_nextstep}

\noindent\textbf{Policies have the skills, yet stall.}\label{sec:exp_stall} Both policies execute the individual steps reliably: on the atomic tasks of RoboCasa365, the WAM succeeds in 79.2\% of episodes and the VLA in 86.2\%. Yet once a long-horizon rollout stalls, it stays stalled: under the task instruction, 97\% of continuations from a stalled state make no further progress within 320 steps. The failure is therefore not a missing skill. The policy holds a skill it does not invoke.

We replace the task instruction with the oracle's next-step instruction and change nothing else. Subgoal completion rises by $4.6\times$ (\cref{fig:stall}a), a gain of 10.5 points (95\% CI [5.3, 17.6]), and the effect holds for both architectures: $4.2\times$ on the WAM and $6.4\times$ on the VLA. The effect is largest where the policy has not started the task, with a $6.3\times$ increase on states stalled before the first subgoal, and it carries through to the end of the task: task success rises by $2.3\times$. The curves keep separating for the full 640 steps.

\noindent\textbf{The gain comes from the content of the instruction.} Two controls on 228 of the stalled states isolate what the instruction contributes (\cref{fig:controls}). A paraphrase of the task instruction leaves subgoal completion unchanged ($+1.3$ points, 95\% CI [$-0.2$, 3.0]), and the next-step instruction exceeds it by 9.4 points [3.3, 18.2]. Naming the wrong step lowers task success below that of the unchanged task instruction, and the correct step exceeds it by $4.6\times$. A stall is resolved by telling the policy the right step, not by telling it something new.

\begin{figure*}[t]
\centering
\includegraphics[width=\textwidth]{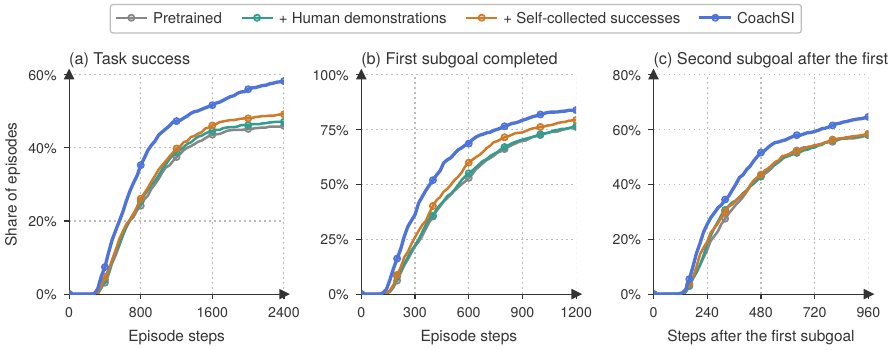}
\caption{\textbf{After \method{}, the policy advances on its own.} Unassisted rollouts on the 18 long-horizon tasks in unseen scenes (1{,}080 episodes per policy). (a) Share of episodes that have succeeded by each step. (b) Share of episodes that have completed a first subgoal. (c) Among episodes that complete a first subgoal, the share that complete the second, by steps after the first.}
\vspace{-5pt}
\label{fig:after}
\end{figure*}

\subsection{A Frozen VLM Is a Sufficient Coach}
\label{sec:exp_vlm}

We replace the oracle with the frozen VLM on the 261 stalled states evaluated with every instruction source (\cref{fig:stall}b). With camera images as its only input, the VLM coach follows the oracle curve over the full horizon and recovers 91\% of the oracle gain at 320 steps (\cref{fig:stall}c); with access to state it recovers all of it. Selecting the next step requires no training, no robot data, and no privileged information: an off-the-shelf VLM already knows it.

\subsection{Coached Experience Becomes Unassisted Capability}
\label{sec:exp_training}
\begin{table}[t]
\centering\small
\setlength{\tabcolsep}{3pt}
\caption{\textbf{Success on long-horizon tasks in unseen scenes} (18 tasks, 1{,}080 episodes). The fine-tuned policies receive the task instruction only; they share the training recipe and the per-task data budget and differ only in the source of the task data. The second row keeps the VLM coach in the loop at test time, without fine-tuning. \method{} is reported for both coaches used in collection. $\Delta$ is the difference to the pretrained policy.}
\label{tab:training}
\begin{tabular}{@{}lccc@{}}
\toprule
Policy & \shortstack{VLM at\\test time} & \shortstack{Success\\(\%)} & $\Delta$\\
\midrule
Pretrained & -- & 47.4 & --\\
Pretrained + VLM coach in the loop & \checkmark & 41.9 & $-5.5$\\
+ Human demonstrations & -- & 49.4 & $+2.0$\\
+ Self-collected successes & -- & 51.9 & $+4.5$\\
\method{} (image-only coach) & -- & 56.8 & $+9.4$\\
\method{} (state-based coach) & -- & \textbf{59.3} & $\mathbf{+11.9}$\\
\bottomrule
\vspace{-15pt}
\end{tabular}
\end{table}

\begin{table}[t]
\centering\small
\caption{\textbf{Ablation of \method{}.} Unassisted success on the 18 long-horizon tasks in unseen scenes (1{,}080 episodes, task instruction only). Each row removes one component and keeps the training recipe and data budget. $\Delta$ is the difference to \method{}.}
\label{tab:ablation}
\begin{tabular}{@{}lcc@{}}
\toprule
Variant & Success (\%) & $\Delta$\\
\midrule
\method{} & \textbf{59.3} & --\\
Without hindsight relabeling & 51.7 & $-7.6$\\
Without the coach (self-collected) & 51.9 & $-7.3$\\
\bottomrule
\vspace{-25pt}
\end{tabular}
\end{table}

\begin{figure}[t]
\centering
\includegraphics[width=\columnwidth]{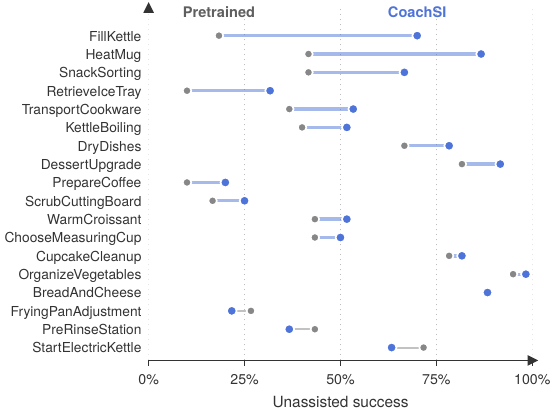}
\caption{\textbf{Unassisted success per task before and after \method{}.} Each row is one long-horizon task, evaluated on 20 unseen scenes with three trials each. \method{} improves on the pretrained policy on 14 of the 18 tasks.}
\vspace{-15pt}
\label{fig:pertask}
\end{figure}

\begin{figure}[t]
\centering
\includegraphics[width=\columnwidth]{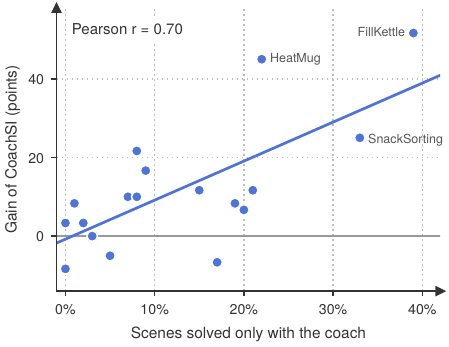}
\caption{\textbf{The gain follows the successes that only coaching produces.} Each point is one task. The horizontal axis is the share of the 100 collection scenes that the policy solves with the coach and fails alone; the vertical axis is the gain in unassisted success of \method{} over the pretrained policy.}
\vspace{-12pt}
\label{fig:scatter}
\end{figure}

\begin{figure*}[!t]
\centering
\includegraphics[width=\textwidth]{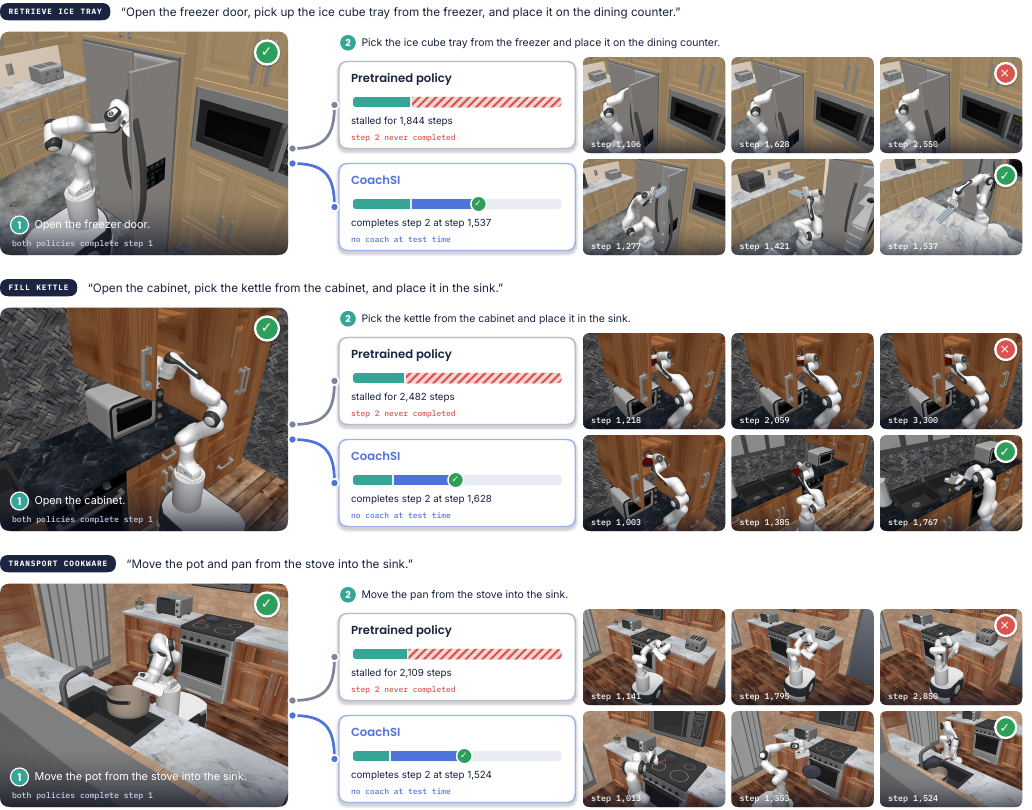}
\caption{\textbf{\method{} continues where the pretrained policy stalls.} Unassisted rollouts on three long-horizon tasks in unseen scenes. In each task, both policies start from the same initial state and receive only the task instruction. \emph{Left:} both policies complete the first step. \emph{Right:} the pretrained policy then stalls and makes no progress for the rest of the episode; \method{} proceeds to the next step and completes the task, with no coach at test time. Bars show each episode on a common step axis: the first step (green), the stall (hatched), and the second step (blue).}
\vspace{-10pt}
\label{fig:qual}
\end{figure*}

\noindent\textbf{Coaching yields successes the policy does not produce alone.} With the image-only coach in the loop, the policy completes 28.7\% of the evaluation episodes it fails on its own. The training data shows the same pattern: we roll out the policy with and without the coach from identical initial states, 100 scenes for each of the 18 tasks, and a third of the 694 coached successes come from initial states where the policy alone fails. These trajectories lie outside what self-collection reaches.

\noindent\textbf{\method{} improves the policy where demonstrations and self-collection do not.} \cref{tab:training} reports unassisted success. Additional human demonstrations raise the pretrained policy from 47.4\% to 49.4\%, and the policy's own successes reach 51.9\%. \method{} reaches 59.3\%, a gain of 11.9 points over the pretrained policy. It exceeds human demonstrations by 9.8 points (95\% CI [4.0, 16.3]) and self-collected successes by 7.3 points [4.0, 10.7], and it does so with fewer trajectories than self-collection (694 against 794). At an equal budget, what matters is not how much the policy practices but whether the practice contains what it cannot yet do.

\noindent\textbf{\method{} works from camera images alone.} With a coach that sees only camera images during collection, \method{} reaches 56.8\% (\cref{tab:training}), 9.4 points above the pretrained policy (95\% CI [2.3, 17.9]), 7.3 above human demonstrations [1.7, 13.9], and 4.8 above self-collected successes [0.6, 9.5].

\noindent\textbf{The coach is better distilled than deployed.} Keeping the coach in the loop at test time, the deployment mode of hierarchical systems~\cite{ahn2022saycan,shi2025hirobot}, yields 41.9\% (\cref{tab:training}): the coach adds successes the policy does not reach alone, but it also redirects episodes the policy would have completed. \method{} keeps only the successes. The fine-tuned policy exceeds the coached pretrained policy by 17.3 points (95\% CI [9.8, 26.5]) and queries no VLM at test time, where the coach in the loop is queried 43 times per episode.

\noindent\textbf{Every component of \method{} contributes.} \cref{tab:ablation} removes one component at a time. Without the coach, the policy trains on its own successes and reaches 51.9\%. Without hindsight relabeling, the coached data keeps the next-step instructions the policy executed, and success falls to 51.7\%, 7.6 points below \method{} (95\% CI [3.8, 11.3]): coached experience becomes a capability under the task instruction because it is relabeled with it.

\noindent\textbf{The policy now takes the next step by itself.} \cref{fig:after} follows the unassisted rollouts over time. By step 1{,}200, \method{} has reached the success rate that the pretrained policy attains only at the end of the episode (\cref{fig:after}a). The difference appears from the first step of the task: within 320 steps of the start, 41.5\% of \method{} episodes have completed a first subgoal, against 23.9\% for the pretrained policy, 24.9\% with human demonstrations, and 28.8\% with self-collected successes (\cref{fig:after}b). After the first subgoal, the policy proceeds to the second more often and sooner (\cref{fig:after}c). On episodes that both policies complete, \method{} finishes in a median of 593 steps, against 735 for the policy fine-tuned on human demonstrations. This is the behavior the coach induced in \cref{fig:stall}, now produced without the coach.

\noindent\textbf{The gain is broad, and the skills were already there.} \method{} improves on the pretrained policy on 14 of the 18 tasks (\cref{fig:pertask}). The largest gains are on tasks the pretrained policy rarely completes: FillKettle rises from 18.3\% to 70.0\% and HeatMug from 41.7\% to 86.7\%. Across tasks, the gain follows the share of scenes that only coaching solves (Pearson $r = 0.70$, $p = 0.001$; \cref{fig:scatter}): the policy improves where the coach showed it successes it could not produce. The action backbone is frozen throughout, so every motor command at test time is produced by the pretrained weights. What changed is the video backbone's prediction of what happens next. \cref{fig:qual} shows the result on three tasks: the pretrained policy completes the first step and stalls for the rest of the episode, while \method{}, from the same initial state and the same task instruction, proceeds to the next step and completes the task. Language coaching taught the policy the next step, and the skills it already had did the rest.

\section{Conclusion}
We presented \method{}, a self-improvement method guided by language alone. A frozen VLM coach tells a stalled policy its next step, and fine-tuning on the relabeled successes turns the coached behavior into the policy's own. A single next-step instruction raises subgoal completion on stalled states by $4.6\times$ on a WAM and a VLA, and an off-the-shelf VLM supplies it from camera images alone. Fine-tuning on the coached experience raises unassisted success in unseen scenes from 47.4\% to 59.3\%, above both human demonstrations and self-collected experience at an equal data budget, with the action backbone frozen. Generalist policies already hold the skills that long-horizon tasks require. What they need is to be told the next step, once.

{\small\bibliographystyle{ieeenat_fullname}\bibliography{main}}
\end{document}